\documentclass[11pt]{article}

\usepackage[margin=1in]{geometry}
\usepackage{lmodern}
\usepackage[T1]{fontenc}
\usepackage[utf8]{inputenc}
\usepackage{amsmath,amssymb}
\usepackage{booktabs}
\usepackage{graphicx}
\usepackage{microtype}
\usepackage{caption}
\usepackage[hidelinks]{hyperref}
\usepackage{natbib}
\newcommand{\mem}{\textsc{mem}}
\newcommand{\gen}{\textsc{gen}}

\title{Beyond Endpoint Scores:\
Time- and Capacity-Conditioned Evaluation of\
Continual Knowledge Updating}

\author{Heejin Choi\\
  Yonsei University \quad Sooil Development \quad NCreate\\
  \texttt{heejin.share@gmail.com}}

\date{}

\begin{document}
\maketitle

\begin{abstract}
Continual knowledge-updating methods are often declared superior from one final
checkpoint and one conventional adapter rank. We show that this evidence can be
insufficient to identify the better operating point. Holding a periodic hierarchy
fixed, we compare it with cumulative replay over a 24-month Wikidata stream while
varying evaluation month, replay LoRA rank, and query formulation. The selected
winner changes across this evaluation region: on Qwen2.5-1.5B (three seeds), the
hierarchy's 5.0-point advantage over rank-8 replay becomes an 11.6-point deficit
against rank-72 replay, and at high ranks a consolidation-aligned endpoint can
suggest a tie while replay leads by 9--13 points on average. The same
rank-conditioned reversal appears on Llama-3.2-1B and persists on held-out
paraphrases.

These findings change what should count as evidence of superiority in continual
updating. A method's ranking is not necessarily a property of a single score; it
can depend on when it is measured and how much replay-side adaptation capacity
the baseline receives. We therefore propose a constructive evaluation protocol:
evaluate throughout the stream, sweep tunable baseline capacity over a
predeclared range, test multiple query formulations, and declare a robust winner
only when the ordering is stable across this region. Otherwise, comparisons
should report winner regions and retention--stability--cost frontiers. Under this
protocol, the periodic hierarchy is a lower-update-cost operating point, not a
quality winner.
\end{abstract}

\section{Introduction}

A deployed LLM's parametric knowledge is fixed at its training cutoff while
the facts users ask about continue to change. Continual pretraining,
parameter-efficient adaptation, model editing, retrieval, and hybrid memory
systems offer different ways to keep a frozen model current. These methods are
often compared by reporting a final score under one selected configuration,
implicitly treating method quality as a scalar.

The broad ingredients of this question have important precedents. Continual
evaluation work shows that boundary-only scores can hide temporary forgetting
and recovery \citep{delange2022}. Knowledge-editor rankings can change with
metric, scoring methodology, and edit batch size \citep{pohl2025principled}.
LoRA-memory studies map rank-dependent factual storage and saturation
\citep{back2026loramemory,mora2024}, while continual-learning work shows that
rank or the trainable parameter regime can change stability--plasticity
trade-offs and even comparative method rankings
\citep{codyra2026,ftregimes2026}. Sequential-editing work likewise separates
immediate acquisition from later accessibility \citep{d4s2024,oneill2026facts}.
We therefore do not claim novelty for any of these individual observations.
Our narrower question is whether \emph{update phase and replay rank jointly}
change the winner of a continual factual-updating comparison when the
hierarchy configuration is held fixed. Within the
public literature covered by our prior-art search through August 2026, we did
not find a study demonstrating this joint hierarchy-versus-replay reversal
across both model family and query surface.

Our results show that this scalar view is inadequate in our setting. The
apparent winner depends on both \textbf{when} the model is evaluated and
\textbf{how much adaptation capacity} the replay baseline receives. At month
23, rank-8 cumulative replay leads a periodic hierarchy by 4.9 points; at
month 24, immediately after consolidation, the hierarchy leads by 13.1
points. Increasing replay rank then moves the entire comparison downward: at
rank 72 the final checkpoint suggests a tie, although time-averaged replay
leads by approximately 9 points on training templates and 13 points on unseen
paraphrases.

The central object of this paper is therefore a \textbf{time--replay-rank
evaluation surface} for a fixed hierarchy configuration:
$\Delta_{t,r,s}=R_{\mathrm{hier}}(t,s)-R_{\mathrm{replay},r}(t,s)$.
Throughout this surface analysis, the hierarchy is fixed at r64 slow + r8 fast
with $K{=}6$; only the evaluation month $t$, replay rank $r$, and query surface
$s$ vary. Replay rank shifts the average level of the comparison, while
periodic consolidation in the fixed hierarchy adds a temporal oscillation.
Selecting one point can therefore exaggerate,
reverse, or hide the average conclusion.

This perspective revises our own initial interpretation. Against conventional
rank-8 replay, the hierarchy appeared both cheaper and better at retention. A
full rank sweep showed that replay was capacity-limited: retention rises
smoothly and saturates near rank 64, and sufficiently capacitated replay
exceeds the hierarchy in both retention and unchanged-fact stability. The
same reversal appears on Qwen2.5-1.5B. The hierarchy's surviving contribution
is a lower-update-cost operating point obtained by replaying the full history
only every $K$ months, not evidence that calendar hierarchy stores knowledge
more effectively.

\paragraph{Contributions.}

\textbf{C1---A joint time--rank winner-reversal analysis.}
Holding the hierarchy configuration fixed (r64 slow + r8 fast, $K{=}6$), we
evaluate the hierarchy-versus-replay margin over update time, replay rank, and
query surface. Consolidation-aligned endpoint bias remains similar across
the tested replay ranks, while rank shifts the average ordering. Their joint
effect yields effect inflation, sign reversal, and hidden quality gaps. This is
the paper's principal novelty claim; the underlying stability-gap and rank
sensitivity phenomena are established prior work.

\textbf{C2---A cross-family falsification of the conventional low-rank
baseline.} The hierarchy appears to win retention against rank-8 replay, but
the ordering reverses against rank-72 replay on Llama-3.2-1B and
Qwen2.5-1.5B, with three seeds and on both training-template and held-out-
paraphrase surfaces. We use the resulting surface to motivate a robust-winner protocol that
predeclares evaluation times, replay capacities, and query surfaces and reports
condition-dependent winner regions when the ordering changes. The rank curve and
elementary replay-cost accounting are supporting analyses, not claims that
rank-dependent LoRA memory or replay complexity are newly discovered.

\textbf{C3---Scoped stream-structured diagnostics.} We report a month-by-month
O-LoRA residual trajectory on a relation-concentrated factual stream and a
controlled WISE contrast in which P54 concentration increases answer
concentration and lowers cumulative accessibility. We explicitly scope these
as empirical diagnostics: orthogonal-capacity exhaustion and immediate-versus-
cumulative degradation have prior art, and the WISE absolute result is limited
to our integration.

\section{Benchmark and evaluation protocol}

\paragraph{Stream.} 24 monthly snapshots (2024-01\,--\,2025-12) of Wikidata
attribute changes with start-time qualifiers; up to 200 changed facts/month
(21 months hit the cap; 198, 194 and 147 in the remainder, so 4{,}739
total); a fame-filtered, end-time-verified \emph{unchanged} control (never
trained on); base-model cutoffs verified to precede the stream.

\paragraph{Scoring.} Length-normalised loglikelihood multiple-choice against
same-relation distractors, per-record. Two surfaces per fact: \mem{} (the
training template) and \gen{} (an unseen paraphrase); every number is
surface-tagged. Retention is mean accuracy over all months seen so far;
stability is accuracy on the unchanged control; both are \textbf{time-averaged
over $t=1\ldots24$} unless explicitly marked as an endpoint reference.

\paragraph{Methods.} B0 frozen; B1 sequential LoRA; B2 full-replay (rank 8 conventional; rank 72 as a nominal-rank
match to the hierarchy's active-rank sum); B3 20\%-replay; B4
O-LoRA \citep{olora2023}; B5 RAG (bge-m3 + FAISS); M1b two-level hierarchy
(r64 slow, $K{=}6$, plus r8 chained fast); M2 three-level; WISE
\citep{wise2024} via EasyEdit, three configurations. Seeds: 3 for
B2r8/B2r72/M1b/M2 and all Qwen/3B runs; 1 for the intermediate sweep points
$r\in\{16,32,64\}$; single-seed results are marked.

\paragraph{Provenance.} Every result JSON records model, method, ranks, seed,
$K$, per-month costs, and a completeness flag; headline tables are regenerated
from the JSONs by script, never hand-edited. All runs use RTX 4090 GPUs with
identical library pins; B2r72 seeds 1--2 and the $r\in\{16,32,64\}$ sweep ran
on rented hosts (device recorded per result; seed 0 local vs seeds 1--2 remote
differ by $\le0.001$ retention, bounding the device effect).

\section{Rankings form a time--replay-rank surface}
\label{sec:timing}

\begin{figure}[t]
\centering
\includegraphics[width=0.86\textwidth]{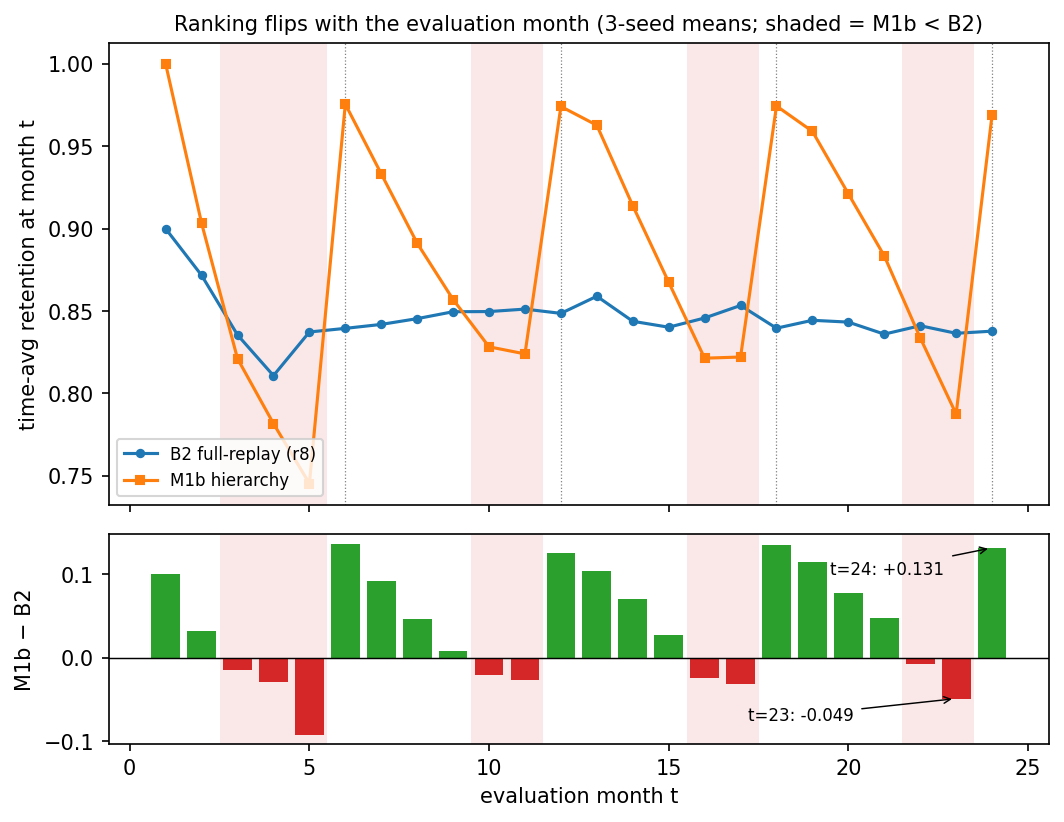}
\caption{The consolidation sawtooth and the months in which the ranking flips
(3-seed means). Shaded months are those where flat replay leads; dotted lines
mark consolidation boundaries. The two candidate endpoints $t{=}23$ and
$t{=}24$ give opposite answers.}
\label{fig:flip}
\end{figure}

Boundary-only evaluation is already known to hide transient loss and recovery
in continual learning \citep{delange2022}. Here we use that concern as the
starting point for a joint capacity test. Consolidation produces a sawtooth:
M1b peaks at $t\in\{6,12,18,24\}$ (\mem{} 0.973 mean at boundaries) and sags
between (0.868 mean mid-window; amplitude $+0.105$, and $+0.135$ on the
paraphrase surface). Because $T{=}24$ is a consolidation boundary, endpoint
evaluation samples the peak (Figure~\ref{fig:flip}). Measured on identical
3-seed runs:

\begin{itemize}
\item \textbf{Ranking flips in 9/24 months} against the conventional rank-8
  baseline---the last 2--3 months of every window. $T{=}23 \Rightarrow$ replay
  $+4.9$pp; $T{=}24 \Rightarrow$ hierarchy $+13.1$pp.
\item \textbf{Effect inflation $3.32\times$}---endpoint margin $+0.131$ vs
  time-averaged $+0.040$.
\item \textbf{Schedule blindness by construction}---$K{=}3/6/12$ all divide
  24; endpoint ${\approx}0.969$ for all three; stream means
  $0.943/0.885/0.813$. This comparison is between variants of one method at
  \emph{identical} capacity, so it isolates the timing effect from any
  capacity confound.
\end{itemize}

\begin{figure}[t]
\centering
\includegraphics[width=0.96\textwidth]{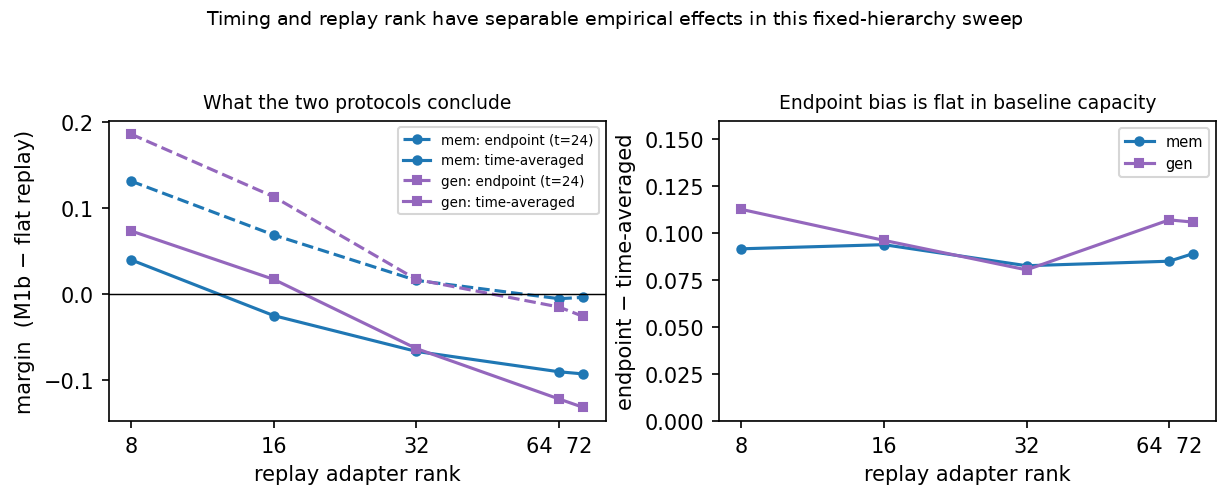}
\caption{For the fixed hierarchy, replay rank changes the method-comparison
margin. Left: endpoint and time-averaged hierarchy-minus-replay margins versus
replay rank. Right: their difference, the endpoint bias. Replay rank shifts the
level, while boundary-aligned timing adds a similar positive bias across the
tested ranks.}
\label{fig:timing}
\end{figure}

\paragraph{Endpoint bias persists across replay capacity.}
A natural objection is that these flips exist only because the rank-8 baseline
is starved (Section~\ref{sec:capacity}). Recomputing the margin against replay
at every swept rank shows otherwise: the \emph{endpoint bias}---endpoint
margin minus time-averaged margin---is essentially constant in rank,
$+0.092/+0.094/+0.083/+0.085/+0.089$ for $r=8/16/32/64/72$ on the
training-template surface ($+0.113$ to $+0.081$ on paraphrases), while the
\emph{level} moves steadily with capacity (months won by the hierarchy:
$15/24 \to 10 \to 6 \to 4 \to 1$). Capacity shifts the average level; boundary-aligned consolidation adds a
similarly sized positive bias across the tested ranks (Figure~\ref{fig:timing}).

Consequently the endpoint distorts the conclusion at every capacity, in three
distinguishable regimes (Table~\ref{tab:regimes}). At intermediate replay ranks
($r{=}16,32$) the endpoint reverses the sign of the conclusion; at saturated
capacity ($r{=}64,72$) it reports a tie where the honest answer is a
nine-point replay win. At every rank the hierarchy's winning months are
consolidation boundaries and their immediate successors---at $r{=}72$ it leads
in exactly one month of 24, and that month is $t{=}6$.

\begin{table}[t]
\centering
\caption{Three regimes of endpoint distortion (\mem{} surface). Positive
margins favour the hierarchy.}
\label{tab:regimes}
\begin{tabular}{rlll}
\toprule
replay rank & endpoint says & time-average says & distortion \\
\midrule
8  & hierarchy $+0.131$ & hierarchy $+0.040$ & $3.3\times$ inflation \\
16 & hierarchy $+0.069$ & \textbf{replay $+0.025$} & \textbf{sign reversal} \\
32 & hierarchy $+0.016$ & \textbf{replay $+0.067$} & \textbf{sign reversal} \\
64 & tie ($-0.006$) & replay $+0.091$ & $16\times$ understatement \\
72 & tie ($-0.004$) & replay $+0.093$ & $23\times$ understatement \\
\bottomrule
\end{tabular}
\end{table}

The demonstrated mechanism is boundary alignment: periodic consolidation
raises retention, and the selected benchmark endpoint happens to sample that
peak. Other methods with periodic merge or consolidation schedules may share
this structural risk, but we demonstrate the quantitative effect only for the
variants evaluated here. Similar concerns about boundary-only reporting have
been raised in continual evaluation \citep{delange2022}.

\section{Rank-conditioned reversal of the method comparison}
\label{sec:capacity}

Prior work already establishes that LoRA's factual-memory capacity increases
and eventually saturates with rank, and that rank controls a stability--
plasticity trade-off \citep{back2026loramemory,mora2024,codyra2026}. We do not
claim those facts as new. Instead, the sweep is a falsification control for our
original method comparison: does the hierarchy still appear preferable when
replay is not fixed at the conventional rank 8?

\begin{figure}[t]
\centering
\includegraphics[width=0.96\textwidth]{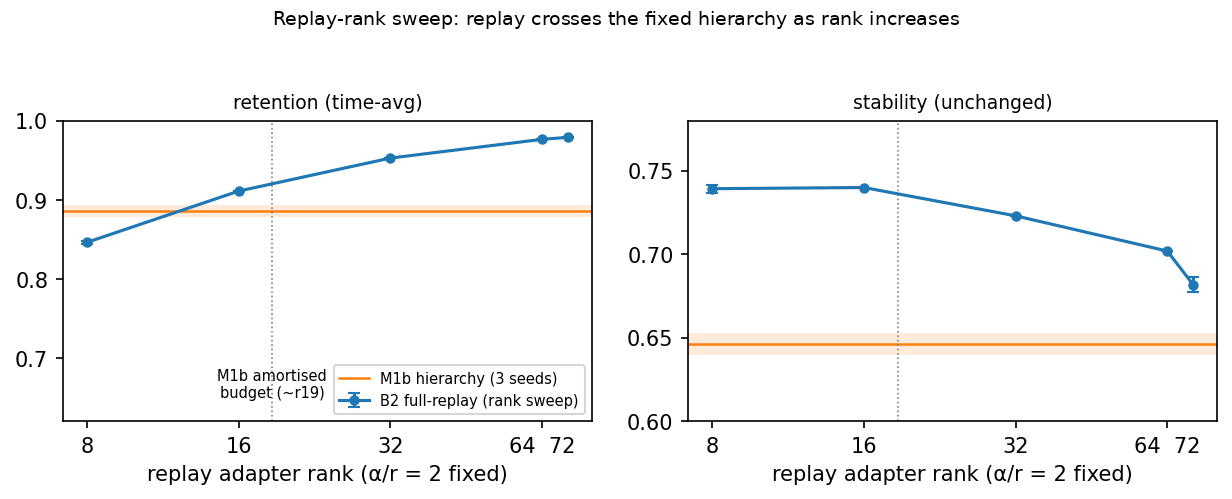}
\caption{Replay-rank sweep used as a falsification control. Consistent with
prior LoRA-memory work, replay retention rises and saturates with rank; the
distinctive result here is that replay crosses the fixed hierarchy. Stability is
flat to rank 16 and then falls. The hierarchy's 3-seed band is shown for
comparison.}
\label{fig:capacity}
\end{figure}

The rank-8 replay baseline places 4{,}739 selected training records into a
low-rank adapter while the hierarchy carries active ranks $64{+}8$. Sweeping
replay rank at fixed $\alpha/r=2$ holds adapter scaling constant while changing
both update rank and trainable parameter count. We call the comparison
\emph{rank-conditioned}, not fully capacity- or compute-matched. Results are
shown in Table~\ref{tab:capacity} and Figure~\ref{fig:capacity}.

\begin{table}[t]
\centering
\caption{Capacity sweep, time-averaged over $t=1\ldots24$. Cost cells are
wall-clock GPU-seconds from the seed-0 runs, which all executed on the same
local RTX 4090; the intermediate sweep points ran on rented hosts and their
timings are \emph{not comparable} (r16 measured 2554\,s against r64's
1983\,s---a lower rank costing 29\% more, which host variation explains and
rank cannot), so we mark them n/c. All accuracy comparisons are
device-independent: the r72 seeds split across local and rented hosts agree to
$\le0.001$ retention.}
\label{tab:capacity}
\begin{tabular}{lccccrc}
\toprule
method & ret (\mem) & ret (\gen) & stab (\mem) & stab (\gen) & GPU-s & seeds \\
\midrule
B2 r8  & $0.846\pm.002$ & $0.741\pm.005$ & $0.739\pm.002$ & $0.713\pm.002$ & 2062 & 3 \\
B2 r16 & 0.911 & 0.797 & 0.740 & 0.716 & n/c & 1 \\
B2 r32 & 0.953 & 0.878 & 0.723 & 0.691 & n/c & 1 \\
B2 r64 & 0.976 & 0.937 & 0.702 & 0.650 & n/c & 1 \\
\textbf{B2 r72} & $\mathbf{0.979\pm.001}$ & $\mathbf{0.947\pm.004}$ & $0.682\pm.005$ & $0.631\pm.003$ & 2230 & 3 \\
M1b (r64+8) & $0.885\pm.006$ & $0.814\pm.008$ & $0.646\pm.005$ & $0.604\pm.005$ & 586 & 3 \\
\bottomrule
\end{tabular}
\end{table}

Three observations hold on both scoring surfaces.
\textbf{(i) The method ordering reverses.} Flat replay beats M1b on
\emph{both} retention and stability from $r{=}16$ on the training-template
surface and from $r{=}32$ on the held-out-paraphrase surface (at $r{=}16$,
\gen{} retention 0.797 still trails M1b's 0.814). The central result is not
that higher rank helps; it is that choosing rank 8 versus a sufficiently
capacious replay point produces opposite conclusions about which updating
strategy is preferable.
\textbf{(ii) The comparison changes systematically over the tested sweep.}
Consistent with prior LoRA-memory work \citep{back2026loramemory}, replay
retention rises and saturates near $r\approx64$ (\mem{}
$0.846\to0.979$; \gen{} $0.741\to0.947$). Ranks 16, 32, and 64 are single-seed
exploratory points; the r8 and r72 endpoints are three-seed results.
\textbf{(iii) Rank is not free.} Stability is unchanged from rank 8 to rank 16
($0.739$ vs $0.740$, a difference below the r8 seed standard deviation) and
then falls steadily (\mem{} $0.723\to0.702\to0.682$; \gen{}
$0.691\to0.650\to0.631$), consistent with prior rank--forgetting analyses
\citep{codyra2026}. We report the crossing points rather than a formal budget
equivalence: LoRA rank is not linearly additive in functional capacity, and
our sweep changes both update rank and parameter count. The hierarchy's robust
advantage is lower update cost ($\sim3.5\times$), not higher quality.

\section{Resource accounting (supporting analysis)}
\label{sec:cost}

The following is elementary accounting included to avoid a misleading
asymptotic claim; it is not a theoretical novelty claim. With period $K$ and
cumulative raw replay, total work is $\sum_i iK \approx T^2/2K + O(T)$; monthly
full replay is $T^2/2$, so the ratio approaches $K$ for fixed $K$ as
$T\to\infty$. Measured ratios are $2.5\times$ at $T{=}12$ (predicted
$2.6\times$) and $3.5\times$ at $T{=}24$ (predicted $3.6\times$), with a
$6\times$ asymptote for $K{=}6$. Thus the method remains quadratic with raw
replay, and bounded active adapter parameters do not imply bounded replay-data
storage. Parameter-only merging \citep{mergebeforeforget2025} targets the
bounded-corpus property that this schedule lacks.

\section{Robustness: cross-model and cross-scale}

Regime-aware continual-learning work shows that changing the trainable
parameter subspace can change comparative method rankings
\citep{ftregimes2026}. Our question is narrower: does a replay-rank-induced
winner reversal in factual updating transfer across LLM family and query
surface?

Against the fixed rank-8 replay baseline, 3 seeds per setting:

\begin{table}[h]
\centering
\begin{tabular}{lccc}
\toprule
setting & M1b ret $-$ B2r8 ret & cost ratio & stability trade \\
\midrule
Llama-3.2-1B  & $+0.040$ & $3.5\times$ & $-0.093$ \\
Qwen2.5-1.5B  & $+0.050$ & $3.5\times$ & $-0.046$ \\
Llama-3.2-3B  & $+0.006$ & $3.2\times$ & $-0.063$ \\
\bottomrule
\end{tabular}
\end{table}

Direction reproduces in all three settings. Within the Llama family the margin
shrinks $6\times$ from 1B to 3B (larger models replay better) while the
hierarchy's stability rises ($0.646\to0.748$), so 1B-class effect sizes should
not be extrapolated upward. The Qwen point ($+0.050$) is a different family
and does not order between them; with three settings across two families we
report the two within-family points as the scale evidence and treat Qwen as a
family-transfer check, not a third point on a trend line.

\paragraph{The capacity reversal is not a Llama-1B artifact.} Every row above
uses the fixed rank-8 baseline that Section~\ref{sec:capacity} shows to be
starved, so on its own the table establishes reproduction of the
\emph{direction}, not of the ranking. We therefore repeated the
rank-conditioned contrast on the second model family
(Table~\ref{tab:qwen}): increasing Qwen's replay rank from 8 to 72 (the hierarchy's nominal active-rank sum) flips the retention margin from $+0.050$ to
$-0.116$ (\mem) and $-0.109$ (\gen), with seed $\sigma\le0.004$. The effect is
in fact stronger than at 1B: on Qwen the rank-72 replay wins
stability as well ($+0.034$), so the hierarchy is dominated on both quality
axes at once, and its surviving advantage is again update cost
($3.24\times$)---the same bound-by-$K$ saving. Intermediate ranks were not
swept on Qwen; this tests family transfer of the reversal, not the shape of
the dose-response.

\begin{table}[h]
\centering
\caption{Rank-conditioned contrast on Qwen2.5-1.5B (3 seeds, time-averaged).
The conventional baseline and the rank-72 baseline give opposite answers.}
\label{tab:qwen}
\begin{tabular}{lcccc}
\toprule
method & ret (\mem) & ret (\gen) & stab (\mem) & GPU-s \\
\midrule
B2 r8 (conventional) & $0.818\pm.002$ & $0.669\pm.019$ & $0.668\pm.001$ & 3874 \\
\textbf{B2 r72} & $\mathbf{0.984\pm.001}$ & $\mathbf{0.868\pm.008}$ & $0.656\pm.004$ & 3603 \\
M1b (r64+8, $K{=}6$) & $0.868\pm.004$ & $0.759\pm.004$ & $0.623\pm.002$ & 1113 \\
\bottomrule
\end{tabular}
\end{table}

The remaining untested cell is a rank-72 replication at 3B.

\section{Stress tests beyond immediate edit success}
\label{sec:homogeneity}

\paragraph{O-LoRA.} O-LoRA constrains each new update toward a subspace
orthogonal to previous updates. Capacity exhaustion in orthogonal-subspace
continual learning is not a new mechanism claim: E$^2$-LoRA explicitly
identifies diffusion across the basis and exhaustion of capacity for future
tasks \citep{e2lora2026}. Our contribution here is a measured trajectory in a
relation-concentrated factual stream. O-LoRA's time-averaged \mem{} retention
is 0.644 and \gen{} retention is 0.594, below plain sequential LoRA's \mem{}
retention of 0.657. From $t{=}2$ through $t{=}23$, the logged residual overlap
$\lVert A_t A_{\mathrm{prev}}^{\top}\rVert_F^2$ increases approximately
linearly from $0.0020$ to $0.0476$, with slope $0.0020$ per prior task and
$R^2{=}0.94$. Month 24 jumps to $0.2017$, but it contains 147 training records
rather than approximately 194--200 and is reported separately. The trajectory
is consistent with progressive subspace pressure in this setting; it does not
establish a universal mechanism or rule out hyperparameter sensitivity.

\paragraph{WISE.} The distinction between immediate editing success and later
accessibility is established prior work \citep{d4s2024,oneill2026facts}.
Moreover, \citet{ragorlearning2026} report a WISE case in which multiple
temporal states collapse to one dominant answer. We therefore do not claim the
first immediate-versus-cumulative gap or the first WISE answer collapse. Our
narrower contribution is a quantitative 200-edit contrast under one
integration. Immediate rewrite accuracy is 1.000 after each edit, while
cumulative multiple-choice retention remains near the frozen base. Holding
code, hyperparameters, edit count, and subject--relation deduplication fixed,
the six-relation arm has top-attractor share 0.585, 45 distinct generated
answers, and cumulative MC 0.390; the P54-only arm has top-attractor share
0.800, 12 distinct answers, and cumulative MC 0.330. P54 concentration thus
increases collapse severity in this contrast. The balanced arm also fails to
exceed the approximately 0.40 frozen-base score, and relation identity is not
fully separated from concentration. We do not infer that homogeneity is the
sole cause of failure or that WISE is generally unusable.

\begin{figure}[t]
\centering
\includegraphics[width=0.62\textwidth]{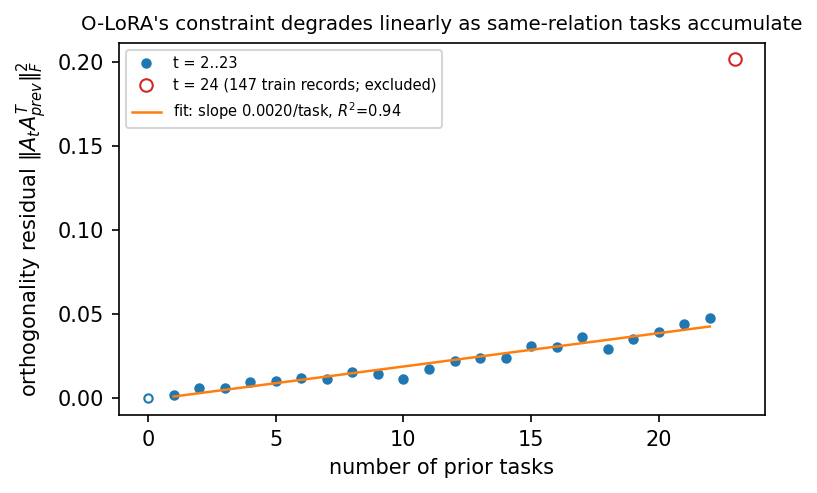}
\caption{O-LoRA's orthogonality residual against prior-task count. The residual
grows approximately linearly through $t{=}23$ ($R^2{=}0.94$); the final month
(open marker) contains fewer training records and is excluded from the fit.}
\label{fig:ortho}
\end{figure}

These stress tests are supporting evidence for cumulative, stream-structured
evaluation rather than the principal novelty of the paper. Operational update
streams can be relation-concentrated, while standard edit benchmarks often use
more diverse objectives; prior sequence-editing work likewise reports effects
of target diversity and sequence length \citep{d4s2024}.

\section{Secondary result: the index-size/retention trade-off in hybrids}
\label{sec:hybrid}

No single store dominates: RAG reaches 0.999 retention at zero training but an
unbounded index ($1{,}955\to11{,}698$) and measurable contamination of
unrelated queries (trivia $-5$pp); rank-72 replay is the best trained quality; the hierarchy is the cheapest. Routing among \{base, adapter, RAG\}
exposes a sharp trade-off rather than a free win:

\begin{table}[h]
\centering
\begin{tabular}{lcc}
\toprule
hybrid variant & retention & index \\
\midrule
spec (old$\to$RAG, recent$\to$adapter) & $0.988\pm.002$ & 11{,}551 \\
bounded (old$\to$adapter, recent$\to$RAG) & $0.746\pm.000$ & \textbf{147} \\
\bottomrule
\end{tabular}
\end{table}

Restricting retrieval to the current consolidation window shrinks the index
$\mathbf{80\times}$ but costs \textbf{24 points of retention}, because the
consolidated adapter---which then must serve the entire past---is the accuracy
bottleneck (it retains only what was injected). The same membership routing
does recover stability at inference ($0.475\to0.623$; 99\% of oracle end-task;
forgetting bounded by router error), and merged adapters serve at base latency
(19.5 vs 21.3\,ms/query). The bounded-index/retention gap is a concrete target
for parameter-level consolidation research.

\section{Related work and novelty boundary}

\paragraph{Continual evaluation and evaluation-dependent rankings.}
Boundary-only evaluation can hide temporary forgetting and recovery;
\citet{delange2022} formalize this stability gap and motivate continual and
worst-case metrics. In knowledge editing, \citet{pohl2025principled} show that
metrics, evaluation methodology, and edit batch size can change editor
rankings. More broadly, \citet{ftregimes2026} show that continual-learning
method rankings need not be invariant to the trainable-depth regime. These
works preclude a broad claim that evaluation-dependent ranking is new. Our
narrower result is a joint update-time by replay-rank analysis in continual
factual updating, with three concrete endpoint error regimes and cross-family
replication.

\paragraph{Rank and parametric knowledge memory.}
\citet{back2026loramemory} systematically map LoRA factual-memory capacity,
rank-dependent saturation, and parameter efficiency on Llama and Qwen. MoRA
shows that low update rank can limit memory-intensive learning under a fixed
parameter budget \citep{mora2024}, and CoDyRA links effective rank to a
stability--plasticity trade-off \citep{codyra2026}. We therefore do not claim
novelty for monotonic rank effects, saturation, or the existence of a rank--
forgetting frontier. The distinct empirical result is that a conventional
rank-8 replay baseline and a higher-rank replay baseline yield opposite
conclusions about hierarchy versus replay, on two model families and two query
surfaces.

\paragraph{Lifelong editing and cumulative accessibility.}
GRACE \citep{grace2023}, WISE \citep{wise2024}, MEMOIR
\citep{memoir2025}, NMKE, and EvoEdit
\citep{nmke2025,evoedit2026} study sequential retention,
generalization, locality, unlearning, and memory organization. D4S documents
sequence-length and target-diversity effects and separates immediate from
cumulative performance \citep{d4s2024}. \citet{oneill2026facts} distinguish
stored content from question-keyed accessibility after later writes, and
\citet{ragorlearning2026} provide a WISE case where temporal states collapse
to one answer. Our WISE analysis is consequently a scoped quantitative stress
test, not a claim to introduce these general failure principles.

\paragraph{Orthogonal subspaces.}
O-LoRA uses orthogonal update subspaces to limit interference
\citep{olora2023}. E$^2$-LoRA explicitly motivates dynamic rank allocation by
the risk that orthogonal methods diffuse energy and exhaust capacity for
future tasks \citep{e2lora2026}. Our residual curve is a new measurement in
this stream, not a new statement of the exhaustion mechanism.

\paragraph{Temporal data and consolidation.}
TemporalWiki \citep{temporalwiki2022} provides an evolving
Wikipedia/Wikidata benchmark, while RAG or Learning? introduces a 2024--2025
continuous-knowledge-drift benchmark \citep{ragorlearning2026}. Recent work
makes the temporal evaluation axis even more explicit: ForgetBench evaluates
forgetting across sequential edit stages \citep{forgetbench2026}, When Does
Continual Learning Require Learning uses a mechanism-agnostic sequential
protocol under environmental drift \citep{whendoes2026}, and OAKS evaluates
online adaptation to evolving knowledge streams \citep{oaks2026}. These works
reinforce that temporal evaluation itself is not our novelty. Progress \&
Compress \citep{progresscompress2018} separates fast acquisition from slow
consolidation; MEMORYLLM, ProCL, Merge-before-Forget, and Memini organize
learned or external memory across updates
\citep{memoryllm2024,procl2026,mergebeforeforget2025,memini2026}. Our benchmark
claim is therefore the joint update-time by replay-rank protocol for a fixed
hierarchy and the resulting winner reversal, not the first temporal factual
stream or the first fast/slow memory architecture.

\section{From winner reversal to a robust evaluation protocol}

\paragraph{Constructive protocol.}
The failure identified here is a model-selection failure, so the immediate
solution is an evaluation protocol rather than a new updater. For a fixed
candidate configuration, predeclare an evaluation region
$\mathcal{E}=\mathcal{T}\times\mathcal{R}\times\mathcal{S}$, where
$\mathcal{T}$ contains evaluation times, $\mathcal{R}$ contains tunable
baseline capacities (replay ranks here), and $\mathcal{S}$ contains query
formulations. At each point report the signed pairwise margin $\Delta(t,r,s)$.
A robust superiority claim requires the ordering to keep the same sign over the
predeclared region, with seed uncertainty reported at headline comparisons. If
the sign changes, the scientifically appropriate result is a winner map or a
retention--stability--cost frontier, not a scalar claim that one method
universally dominates the other.

In practice, this means: evaluate throughout the stream rather than only at the
final checkpoint; disclose whether reported endpoints align with reset, merge,
replay, or consolidation events; sweep or explicitly condition on the capacity
of tunable baselines; report exact trainable parameters and effective update
rank when available; test both training formulations and held-out query
formulations; and account separately for retention, unchanged-fact stability,
update work, replay/index storage, and serving cost. We deliberately do not
collapse this surface into a new single metric, because deployment preferences
over time, quality, and resource cost differ. This protocol resolves the
immediate decision problem exposed by our experiments; designing an updater
that approaches high-rank replay quality at periodic-consolidation cost is a
separate algorithmic target.

\paragraph{Limitations.} The central claim is an empirical
winner reversal within the evaluated time, rank, model, and query ranges; it
is not a universal ordering of updating methods. The complete rank
dose-response is measured on Llama-3.2-1B. Its r8 and r72 endpoints are
three-seed results and replicate on Qwen2.5-1.5B, but the intermediate Llama
ranks are single-seed, intermediate Qwen ranks are untested, and rank-72 replay
at 3B remains untested. Varying LoRA rank changes both trainable parameter
count and effective update rank, so rank 72 is a nominal comparison point, not
a formal capacity- or compute-equivalence to the hierarchy. The stream is one
domain and is dominated by P54; the WISE relation contrast does not fully
match answer entropy, relation difficulty, or target perplexity, and other
editing studies report different diversity effects in other regimes. The
stream does not implement full versioned semantics for repeated changes to the
same subject--relation pair. \gen{} is an unseen paraphrase, not unrestricted
free generation or multi-hop use. WISE's absolute performance is scoped to our
EasyEdit integration, and the O-LoRA final-month spike is confounded with a
smaller training batch. The $O(T^2/K+T)$ expression is transparent resource
accounting, not a theoretical contribution. Finally, while our search through
September 3, 2026 found no prior public study with the same joint time--rank
hierarchy-versus-replay reversal, literature absence cannot be proven and
concurrent work may further narrow the novelty boundary.

\paragraph{Conclusion.} Prior work already establishes that boundary-only
evaluation can hide transient instability, that evaluation protocols can
change editor rankings, and that LoRA rank controls factual-memory capacity.
Our result is their joint consequence for method selection: with the hierarchy
configuration fixed, update phase and replay rank are large enough to change
the hierarchy-versus-replay winner in continual factual updating. A single
endpoint at a single default rank can inflate a small advantage, reverse the
apparent ordering, or hide a large replay lead. The reversal appears on
held-out paraphrases and on a second model family. The solution offered by this
paper is a robust evaluation protocol: claim a winner only when the ordering
survives the predeclared evaluation region; otherwise report the winner regions
and Pareto frontier. Under this protocol, the periodic hierarchy is a
lower-update-cost operating point rather than a quality winner.

\paragraph{Availability.} We plan to release the benchmark stream, per-run
result JSONs with provenance manifests, and all analysis and figure scripts in
a public repository accompanying the preprint.

\bibliography{refs}

@inproceedings{wise2024,
  title     = {{WISE}: Rethinking the Knowledge Memory for Lifelong Model
               Editing of Large Language Models},
  author    = {Wang, Peng and Li, Zexi and Zhang, Ningyu and Xu, Ziwen and
               Yao, Yunzhi and Jiang, Yong and Xie, Pengjun and Huang, Fei and
               Chen, Huajun},
  booktitle = {Advances in Neural Information Processing Systems (NeurIPS)},
  volume    = {37},
  pages     = {53764--53797},
  year      = {2024},
  note      = {arXiv:2405.14768},
}

@inproceedings{progresscompress2018,
  title     = {Progress \& Compress: A Scalable Framework for Continual
               Learning},
  author    = {Schwarz, Jonathan and Czarnecki, Wojciech and
               Luketina, Jelena and Grabska-Barwinska, Agnieszka and
               Teh, Yee Whye and Pascanu, Razvan and Hadsell, Raia},
  booktitle = {Proceedings of the 35th International Conference on Machine
               Learning (ICML)},
  volume    = {80},
  year      = {2018},
}

@inproceedings{temporalwiki2022,
  title     = {{TemporalWiki}: A Lifelong Benchmark for Training and
               Evaluating Ever-Evolving Language Models},
  author    = {Jang, Joel and Ye, Seonghyeon and Lee, Changho and
               Yang, Sohee and Shin, Joongbo and Han, Janghoon and
               Kim, Gyeonghun and Seo, Minjoon},
  booktitle = {Proceedings of the 2022 Conference on Empirical Methods in
               Natural Language Processing (EMNLP)},
  year      = {2022},
  note      = {arXiv:2204.14211},
}

@inproceedings{grace2023,
  title     = {Aging with {GRACE}: Lifelong Model Editing with Discrete
               Key-Value Adaptors},
  author    = {Hartvigsen, Thomas and Sankaranarayanan, Swami and
               Palangi, Hamid and Kim, Yoon and Ghassemi, Marzyeh},
  booktitle = {Advances in Neural Information Processing Systems (NeurIPS)},
  year      = {2023},
  note      = {arXiv:2211.11031},
}

@inproceedings{memoryllm2024,
  title     = {{MEMORYLLM}: Towards Self-Updatable Large Language Models},
  author    = {Wang, Yu and Gao, Yifan and Chen, Xiusi and Jiang, Haoming and
               Li, Shiyang and Yang, Jingfeng and Yin, Qingyu and Li, Zheng and
               Li, Xian and Yin, Bing and Shang, Jingbo and McAuley, Julian},
  booktitle = {Proceedings of the 41st International Conference on Machine
               Learning (ICML)},
  year      = {2024},
  note      = {arXiv:2402.04624},
}

@inproceedings{memoir2025,
  title     = {{MEMOIR}: Lifelong Model Editing with Minimal Overwrite and
               Informed Retention for {LLM}s},
  author    = {Wang, Ke and Qin, Yiming and Dimitriadis, Nikolaos and
               Favero, Alessandro and Frossard, Pascal},
  booktitle = {Advances in Neural Information Processing Systems (NeurIPS)},
  year      = {2025},
  note      = {arXiv:2506.07899},
}

@article{mergebeforeforget2025,
  title   = {Merge before Forget: A Single {LoRA} Continual Learning via
             Continual Merging},
  author  = {Qiao, Fuli and Mahdavi, Mehrdad},
  journal = {arXiv preprint arXiv:2512.23017},
  year    = {2025},
}

@article{procl2026,
  title   = {Continual Fine-Tuning of Large Language Models via Program Memory},
  author  = {Le, Hung and Venkatesh, Svetha},
  journal = {arXiv preprint arXiv:2605.13162},
  year    = {2026},
}

@inproceedings{memini2026,
  title     = {Continual Knowledge Updating in {LLM} Systems: Learning Through
               Multi-Timescale Memory Dynamics},
  author    = {Pattichis, Andreas and Dovrolis, Constantine},
  booktitle = {ICML 2026 Workshop on Continual Adaptation at Scale: Towards
               Sustainable AI (CATS)},
  year      = {2026},
  note      = {arXiv:2605.05097; poster},
}

@inproceedings{olora2023,
  title     = {Orthogonal Subspace Learning for Language Model Continual
               Learning},
  author    = {Wang, Xiao and Chen, Tianze and Ge, Qiming and Xia, Han and
               Bao, Rong and Zheng, Rui and Zhang, Qi and Gui, Tao and
               Huang, Xuanjing},
  booktitle = {Findings of the Association for Computational Linguistics:
               EMNLP},
  year      = {2023},
  note      = {arXiv:2310.14152},
}

@inproceedings{nmke2025,
  title     = {Edit Less, Achieve More: Dynamic Sparse Neuron Masking for
               Lifelong Knowledge Editing in {LLM}s},
  author    = {Liu, Jinzhe and Sun, Junshu and Shen, Shufan and Yang, Chenxue
               and Wang, Shuhui},
  booktitle = {Advances in Neural Information Processing Systems (NeurIPS)},
  year      = {2025},
  note      = {arXiv:2510.22139},
}

@inproceedings{evoedit2026,
  title     = {{EvoEdit}: Evolving Null-Space Alignment for Robust and
               Efficient Knowledge Editing},
  author    = {Lyu, Sicheng and Gu, Yu and Wang, Xinyu and Huang, Jerry and
               Luan, Sitao and Cui, Yufei and Chang, Xiao-Wen and Lu, Peng},
  booktitle = {Findings of the Association for Computational Linguistics: ACL},
  year      = {2026},
  note      = {arXiv:2510.13851},
}

@inproceedings{delange2022,
  title     = {Continual Evaluation for Lifelong Learning: Identifying the
               Stability Gap},
  author    = {De Lange, Matthias and van de Ven, Gido M. and Tuytelaars, Tinne},
  booktitle = {International Conference on Learning Representations (ICLR)},
  year      = {2023},
  note      = {arXiv:2205.13452},
}

@article{mora2024,
  title   = {{MoRA}: High-Rank Updating for Parameter-Efficient Fine-Tuning},
  author  = {Jiang, Ting and Huang, Shaohan and Luo, Shengyue and Zhang, Zihan
             and Huang, Haizhen and Wei, Furu and Deng, Weiwei and Sun, Feng
             and Zhang, Qi and Wang, Deqing and Zhuang, Fuzhen},
  journal = {arXiv preprint arXiv:2405.12130},
  year    = {2024},
}

@inproceedings{d4s2024,
  title     = {Reasons and Solutions for the Decline in Model Performance
               after Editing},
  author    = {Huang, Xiusheng and Liu, Jiaxiang and Wang, Yequan and Liu, Kang},
  booktitle = {Advances in Neural Information Processing Systems (NeurIPS)},
  year      = {2024},
}

@inproceedings{pohl2025principled,
  title     = {Towards a Principled Evaluation of Knowledge Editors},
  author    = {Pohl, Sebastian and Ploner, Max and Akbik, Alan},
  booktitle = {Proceedings of the First Workshop on Large Language Model
               Memorization (L2M2)},
  pages     = {47--60},
  publisher = {Association for Computational Linguistics},
  year      = {2025},
  doi       = {10.18653/v1/2025.l2m2-1.4},
}

@inproceedings{back2026loramemory,
  title     = {Understanding {LoRA} as Knowledge Memory: An Empirical Analysis},
  author    = {Back, Seungju and Lee, Dongwoo and Kang, Naun and Lee, Taehee
               and Hong, S. K. and Gwon, Youngjune and Ahn, Sungjin},
  booktitle = {International Conference on Machine Learning (ICML)},
  year      = {2026},
  note      = {arXiv:2603.01097},
}

@article{ftregimes2026,
  title   = {Fine-Tuning Regimes Define Distinct Continual Learning Problems},
  author  = {Iordache, Paul-Tiberiu and Burceanu, Elena},
  journal = {arXiv preprint arXiv:2604.21927},
  year    = {2026},
}

@article{codyra2026,
  title   = {Take Only What You Need: Rank Minimization as an Implicit
             Forgetting Regularizer in Continual Learning},
  author  = {Lu, Haodong and Zhao, Chongyang and Xue, Jason and Yao, Lina and
             Moore, Kristen and Gong, Dong},
  journal = {arXiv preprint arXiv:2412.01004},
  year    = {2026},
  note    = {First public version 2024},
}

@inproceedings{e2lora2026,
  title     = {Energy-Structured Low-Rank Adaptation for Continual Learning},
  author    = {Li, Longhua and Qi, Lei and Tian, Qi and Geng, Xin},
  booktitle = {International Conference on Machine Learning (ICML)},
  year      = {2026},
  note      = {arXiv:2605.27482},
}

@article{ragorlearning2026,
  title   = {{RAG} or Learning? Understanding the Limits of {LLM} Adaptation
             under Continuous Knowledge Drift in the Real World},
  author  = {Liu, Hanbing and Cao, Lang and Li, Yang},
  journal = {arXiv preprint arXiv:2604.05096},
  year    = {2026},
}

@article{oneill2026facts,
  title   = {Can a Language Model Learn Facts Continually in Its Weights?},
  author  = {O'Neill, Charles},
  journal = {arXiv preprint arXiv:2607.11020},
  year    = {2026},
}

@article{forgetbench2026,
  title   = {{ForgetBench}: Benchmarking Forgetting Dynamics of Long-Term
             Parametric Memory in Language Models},
  author  = {Gu, Ruxi and Zhang, Zhenliang and Wang, Wei},
  journal = {arXiv preprint arXiv:2607.26455},
  year    = {2026},
}

@article{whendoes2026,
  title   = {When Does Continual Learning Require Learning},
  author  = {Harrington, Anne and Saxena, Nayan and Murphy, Michael and
             Borovykh, Anastasia and Yun, Zeyu and Kamath, Sridhar and
             Kyi, Ara Eindra and Darrell, Trevor and Malik, Jitendra and Bai, Yutong},
  journal = {arXiv preprint arXiv:2607.07847},
  year    = {2026},
}

@article{oaks2026,
  title   = {Can Large Language Models Keep Up? Benchmarking Online Adaptation
             to Continual Knowledge Streams},
  author  = {Kim, Jiyeon and Lee, Hyunji and Zhou, Dylan and Park, Sue Hyun and
             Yoon, Seunghyun and Bui, Trung and Dernoncourt, Franck and
             Cha, Sungmin and Seo, Minjoon},
  journal = {arXiv preprint arXiv:2603.07392},
  year    = {2026},
}

\end{document}